\documentclass[journal]{IEEEtran}
\usepackage[pdftex]{graphicx}
\usepackage{algorithm}
\usepackage{algorithmic}
\usepackage[numbers,sort&compress,square]{natbib}
\usepackage{enumerate}
\usepackage{setspace}
\usepackage{caption}
\usepackage{subcaption}
\usepackage{amsmath,amsfonts,amssymb}
\usepackage{multirow}
\usepackage{float}
\usepackage{placeins}
\usepackage{caption}
\usepackage{subcaption}
\usepackage{xcolor}

\ifCLASSINFOpdf
\else
\fi
\usepackage{url}
\begin{document}
\title{Multi-Domain Adversarial Feature Generalization for Person Re-Identification}

\author{Shan~Lin,~\IEEEmembership{Student Member,~IEEE,}
        Chang-Tsun~Li,~\IEEEmembership{Senior Member,~IEEE,}
        and~Alex C.~Kot,~\IEEEmembership{Fellow,~IEEE}% <-this % stops a space
% \thanks{This work is supported by EU Horizon 2020 project, entitled Computer Vision Enable Multimedia Forensics and People Identification (acronym:IDENTITY, Project ID:690907)}% <-this % stops a space
% \thanks{Dataset was collected at the Nanyang
% Technological University, Singapore with the help of the Rapid-Rich Object Search (ROSE) Lab and Smart Mobility Experience Laboratory}% <-this % stops a space
% \thanks{Manuscript received April 19, 2005; revised August 26, 2015.}
\thanks{This work was carried out at the Rapid-Rich Object Search (ROSE) Lab, Nanyang Technological University, Singapore. The research is supported by the EU Horizon 2020 Marie Sklodowska-Curie Actions through the project entitled Computer Vision Enabled Multimedia Forensics and People Identification (Project No. 690907, Acronym: IDENTITY) and the National Research Foundation, Singapore under its AI Singapore Programme (AISG Award No: AISG-100E-2018-018) with the Defence Science and Technology Agency, Singapore, under the project agreement No. DST000ECI19300431. Any opinions, findings and conclusions or recommendations expressed in this material are those of the author(s) and do not reflect the views of National Research Foundation, Singapore.}
}


\markboth{IEEE TRANSACTIONS ON IMAGE PROCESSING, April~2020}%
{Shell \MakeLowercase{\textit{et al.}}: Bare Demo of IEEEtran.cls for IEEE Journals}



\maketitle

\begin{abstract}
With the assistance of sophisticated training methods applied to single labeled datasets, the performance of fully-supervised person re-identification (Person Re-ID) has been improved significantly in recent years. However, these models trained on a single dataset usually suffer from considerable performance degradation when applied to videos of a different camera network. To make Person Re-ID systems more practical and scalable, several cross-dataset domain adaptation methods have been proposed, which achieve high performance without the labeled data from the target domain. However, these approaches still require the unlabeled data of the target domain during the training process, making them impractical. A practical Person Re-ID system pre-trained on other datasets should start running immediately after deployment on a new site without having to wait until sufficient images or videos are collected and the pre-trained model is tuned. To serve this purpose, in this paper, we reformulate person re-identification as a multi-dataset domain generalization problem. We propose a multi-dataset feature generalization network (MMFA-AAE), which is capable of learning a universal domain-invariant feature representation from multiple labeled datasets and generalizing it to `unseen' camera systems. The network is based on an adversarial auto-encoder to learn a generalized domain-invariant latent feature representation with the Maximum Mean Discrepancy (MMD) measure to align the distributions across multiple domains. Extensive experiments demonstrate the effectiveness of the proposed method. Our MMFA-AAE approach not only outperforms most of the domain generalization Person Re-ID methods, but also surpasses many state-of-the-art supervised methods and unsupervised domain adaptation methods by a large margin.
\end{abstract}


% \fi
\IEEEpeerreviewmaketitle

\section{Introduction}
Re-identifying a person in CCTV surveillance systems, also known as Person Re-ID, is a critical but also labor-intensive task. In recent years, the computer vision community has proposed various methods to automatically identify re-appearing people in multi-camera surveillance systems. Most of these proposed approaches are modeled, trained, and tested on the same dataset collected from a very small camera network \cite{GOG,SpindleNet,TriNet,AlignedReID,MGN,BFE,Wu2018What-and-whereRe-identification,Wu2020DeepRe-Identification}. However, a real-world CCTV system usually consists of tens to hundreds of cameras. Detecting, extracting and annotating thousands of people across hundreds of cameras is an extremely challenging task. Hence, using the actual annotated data collected from every target surveillance camera to train a fully-supervised Person Re-ID model is not a practical approach. Besides, most conventional supervised single-dataset models often over-fit to specific datasets (camera networks). Once these supervised models are trained on a given dataset, they usually suffer from considerable performance degradation when applied to a different camera network. Table \ref{tab:performance_degradation} illustrates the performance of a simple supervised model tested on the same dataset and a different dataset. This model uses the ResNet50 network \cite{ResNet} as the feature extraction backbone. In Table \ref{tab:performance_degradation}, the model trained on the Market-1501 dataset \cite{Market-1501} can achieve 91.6 \% Rank 1 accuracy with 78.7\% mAP score when tested on the same dataset. However, it can only achieve 37.6\% Rank 1 and 22.6\% mAP when tested on the DukeMTMC-reID dataset \cite{DukeMTMC-reID}. The performance of the model trained on the DukeMTMC-reID dataset also drops from 83.4\% Rank-1 with 66.6\% mAP to only 48.2\% Rank 1 with 21.6\% mAP when tested on the Market-1501 dataset. This suggests that models trained on a single dataset are prone to over-fitting and have poor generalization performance.

% \begin{table}[h]
% \centering
% \caption{The performance degradation of the single-dataset-trained baseline model (ResNet50) when tested on a different dataset}
% \resizebox{0.48\textwidth}{!}{%
% \begin{tabular}{c|c|c|c|c}
% \hline
% \multirow{2}{*}{Training Dataset} & \multicolumn{2}{c|}{Market1501} & \multicolumn{2}{c|}{DukeMTMC-reID} \\ \cline{2-5} 
%  & Rank 1 & mAP & Rank 1 & mAP \\ \hline
% Market-1501 & 91.6\% & 78.7\% & 37.6\% & 22.6\% \\ \hline
% DukeMTMC-reID & 48.2 \% & 21.6\% & 83.4\% & 66.6\% \\ \hline
% \end{tabular}}
% \label{tab:performance_degradation}
% \end{table}

\begin{table}[h]
\centering
\caption{The performance degradation of the single-dataset-trained baseline model (ResNet50) when tested on a different dataset.}
\resizebox{0.48\textwidth}{!}{%
\begin{tabular}{l|c|c|c|c}
\hline
\multirow{3}{*}{\textbf{Training Dataset}} & \multicolumn{4}{c|}{\textbf{Testing Dataset}} \\ \cline{2-5} 
 & \multicolumn{2}{c|}{Market1501} & \multicolumn{2}{c|}{DukeMTMC-reID} \\ \cline{2-5} 
 & Rank 1 & mAP & Rank 1 & mAP \\ \hline
\multicolumn{1}{c|}{Market-1501} & 91.6\% & 78.7\% & 37.6\% & 22.6\% \\ \hline
\multicolumn{1}{c|}{DukeMTMC-reID} & 48.2 \% & 21.6\% & 83.4\% & 66.6\% \\ \hline
\end{tabular}}
\label{tab:performance_degradation}
\end{table}

The weak generalization capacity and poor scalability in most single-dataset-trained models severely hinder the real-world deployment of Person Re-ID systems. Different datasets are often collected in very different environments (e.g., indoors/outdoors, summer/winter, daytime/nighttime). If we consider each dataset (camera system) as a domain, there are often large domain gaps between datasets. Hence, recent researches focus on unsupervised cross-dataset \textit{domain adaptation} (DA) for Person Re-ID \cite{SPGAN,TJ-AIDL,MMFA,HHL} to obviate the need for annotating the images from new camera systems. These cross-dataset DA methods aim to adapt a model trained on an annotated source dataset to an unlabeled target dataset by image translation, feature alignment, or multi-task learning. By transferring the domain-specific knowledge, the cross-dataset domain adaptation methods do not require labeled (i.e., annotated) data of the target domain. However, for the DA approaches to be effective, the following two issues are yet to be resolved.
\begin{enumerate}
\item \textit{Generalization Issue}: Cross-dataset domain adaptation methods require a large amount of unlabeled data from the target network prior to model adaptation training. However, it may not be known in advance where the model would be deployed. The unlabeled data collection also takes time even if the target site is known, especially when images/videos of all four seasons are required. The additional data collection process will delay the system deployment. 
\item \textit{Scalability Issue}: Cross-dataset domain adaptation methods require the training and fine-tuning of a bespoke model for every new camera network. The training or fine-tuning for a new model may take from hours to days, depending on the system scale. Besides, the scales and configurations of CCTV systems may not always be constant. More new cameras may be added to the system to meet the ever-changing demands. Such changes require the model to be re-trained in order to accommodate the new cameras.
\end{enumerate}

In this paper, we address the generalization and scalability issues of Person Re-ID from a different perspective. Since no single dataset can cover all possible backgrounds and imaging conditions, we decide to learn a universal feature representation from multiple datasets. In recent years, many large-scale Person Re-ID datasets such as CUHK02 \cite{CUHK02}, CUHK03 \cite{CUHK03}, Market-1501 \cite{Market-1501}, DukeMTMC-reID \cite{DukeMTMC-reID}, MSMT17 \cite{MSMT17}, RAP \cite{RAP}, and CUHK-SYSU \cite{CUHK-SYSU} have been collected. They cover a wide variety of visual scenes with various camera settings. Each dataset can be considered as a different surveillance system representing a different domain. Therefore, we reformulate Person Re-ID as a \textit{domain generalization} (DG) problem, in which we train a model from multiple existing datasets without any prior knowledge of the target system (i.e., no domain adaptation). We aim to develop a domain generalization model that can leverage the labeled images from multiple datasets to learn a domain-invariant feature representation. Domain generalization applied to the feature learning on these datasets helps learn a representation that can be relatively well generalized to any unseen surveillance system. This setting simulates the real-world scenario, in which a strong feature learner only needs to be trained on multiple datasets once and can be deployed to new camera networks without further data collection or adaptation training. 
 
%  In the multi-datasets feature generalization experiments, the Person Re-ID models can only be trained from several labelled source datasets, without any domain adaptation process. The testing for these models is performed on four novel unseen datasets without any cross-dataset adaptation process. 

However, due to its challenging nature, few methods have attempted the domain generalization setting \cite{DGD,DIMN,DualNorm}. The recent DIMN \cite{DIMN} sets a standard training and evaluation procedure for the multi-dataset domain generalization for Person Re-ID. The DIMM method is based on a complicated meta-learning procedure. However, the dynamic model synthesis during the testing process makes the DIMN model relatively slow and cumbersome. The DualNorm method \cite{DualNorm} uses a domain style normalization by performing instance normalization (IN) in the early layers of the feature extractor networks such as MobileNet and ResNet. The DualNorm method is efficient and can be integrated into most of the existing Person Re-ID methods. However, it does not fully utilize the domain label for training. 

In this paper, we proposed a novel framework for domain generalization, which aims to learn a universal representation via domain-based adversarial learning while aligning the distribution of mid-level features between them. Our proposed framework can be considered as an extension of our \textbf{M}ulti-task \textbf{M}id-level \textbf{F}eature \textbf{A}lignment (MMFA) network \cite{MMFA} in a multiple domain learning setting. We called it MMFA with \textbf{A}dversarial \textbf{A}uto-\textbf{E}ncoder (MMFA-AAE). Our MMFA-AAE can simultaneously minimize the losses of data reconstruction, identity, and triplet loss. It alleviates the domain difference via adversarial training and also matches the distribution of mid-level features across multiple datasets. Our contributions can be summarized as follows. 
\begin{enumerate}
\item We propose an effective feature generalization mechanism utilizing domain-based adversarial learning. We introduce an additional feature distribution alignment (i.e., Maximum Mean Discrepancy \cite{MMD}) to regularize the feature learning process. By integrating the adversarial auto-encoder \cite{AAE} and Maximum Mean Discrepancy (MMD) alignment, our MMFA-AAE architecture is capable of extracting domain-invariant features from multiple source datasets and generalize the features to unseen target domains (datasets).
\item The proposed MMFA-AAE method not only demonstrates the state-of-the-art performance on the multi-dataset domain generalization setting but also surpasses many domain adaptation Person Re-ID methods. 
\item Unlike the DIMN \cite{DIMN} and DualNorm methods \cite{DualNorm}, our MMFA-AAE reduces the dimension of the feature vectors to only 512 without affecting the overall performance. It can significantly shorten the subject retrieval time and reduce the storage requirement for saving the processed features.
\item Our domain-based adversarial learning sub-network can be easily integrated into most existing Person Re-ID methods. It can help to boost the generalization capacity of the existing Person Re-ID models.
\end{enumerate}

\section{Related Work}
\subsection{Single-dataset Person Re-ID}
In recent years, Person Re-ID methods are often based on deep convolutional neural networks. Early deep learning based approaches are developed based the  Siamese architecture \cite{CUHK03,Huang2019Multi-PseudoRe-Identification,Wu2018What-and-whereRe-identification, Lin2017End-to-EndRe-Identification} to learn the the corresponding regions matching between two input images. The recent methods \cite{Ensemble,MGN,StrongBaseline} are usually consisted of both softmax classification loss and triplet verification loss. The latest approaches, such as DCC \cite{Wu2020DeepRe-Identification} and DuATM \cite{Si2018DualRe-identification}, utilize the attention mechanism to further boost the Person Re-ID performance. Most of these methods are trained and tested on a single dataset to evaluate their performance. However, different datasets are collected from different cameras under different imaging conditions. It has been noted that these supervised single-dataset methods often over-fit to the training dataset and generalize poorly when tested on other unseen datasets. Collecting a labeled dataset for a new camera system is an expensive and time-consuming task. Hence, many recent methods are focusing on cross-dataset domain adaptation (DA) learning \cite{DTR,AdaRSVM,UMDL,SPGAN,TJ-AIDL,MMFA,HHL}.

\subsection{Cross-dataset Domain Adaption Person Re-ID}
DA approaches assume that there exists a massive amount of unlabeled data obtained from the target camera system (also known as the \textit{target domain}). These approaches utilize the information extracted from the unlabeled target domain data to help the models trained on the source domain to adapt to the target domain. Early proposed cross-dataset DA approaches rely on weak label information in the target dataset \cite{DTR,AdaRSVM}. Therefore, these methods can only be considered as semi-supervised or weakly-supervised learning. Recent cross-dataset works, such as UMDL \cite{UMDL}, SPGAN \cite{SPGAN}, TJ-AIDL \cite{TJ-AIDL}, MMFA \cite{MMFA}, do not require any labeled information from the target dataset and can be considered as fully unsupervised cross-dataset domain adaptation learning. The UMDL method tries to transfer the view-invariant feature representation via multi-task dictionary learning on both the source dataset and the target dataset. The SPGAN approach uses the generative adversarial network (GAN) to generate a new training dataset by transferring the image style from the target dataset to the source dataset while preserving the source identity information. Hence, the supervised training on the transferred dataset can automatically adapt to the target domain. The TJ-AIDL approach individually trains two models: an identity classification model and an attribute recognition model. Domain adaptation in TJ-AIDL is achieved by minimizing the distance between the inferred attributes from the identity classification model and the predicted attributes from the attribute recognition model. Similar to TJ-AIDL, the MMFA network is jointly optimized through people identity classification and attribute learning with cross-dataset mid-level feature alignment regularization. In this way, the learned feature representation can be better generalized from one dataset to another. In \cite{HHL}, the Hetero-Homogeneous Learning (HHL) method improves the capability of generalization to target datasets by achieving camera invariance and domain connectedness simultaneously. The BUC \cite{BUC} and the PurifyNet \cite{PurifyNet}, on the other hand, try to estimate labels for the target domain dataset. Compared to previous unsupervised single dataset approaches, recent unsupervised cross-dataset domain adaptation methods yield much better performance. Although DA approaches do not require labeled data from the target domain, they do require a large amount of unlabeled data from the target domain to facilitate the adaptation. They require additional time for data collection and model adaptation, which will delay the system deployment. Compared to DA methods, domain generalization (DG) approaches are relatively more practical in real-world applications.

\subsection{Multi-dataset Domain Generalization Person Re-ID}
Multi-dataset domain generalization (DG) methods aim to learn a universal domain-invariant feature representation that is robust to various domain-shift across different datasets (camera systems). As a result, a domain generalization model can be incorporated into a new surveillance system without fine-tuning and adaptation. In the Person Re-ID research community, only a few works focus on multi-domain generalization \cite{DGD,DIMN,DualNorm}. The Domain Guided Dropout (DGD) method \cite{DGD} is the first multi-dataset domain generalization work. By removing the domain-specific neurons, the DGD method achieves multi-domain generalization by only utilizing the neurons that are effective across all domains. The DGD method is only trained on several small Person Re-ID datasets such as CUHK01 \cite{CUHK01} and CUHK03 \cite{CUHK03}. It only performs the evaluation on the same dataset without considering the cross-dataset situation. DIMN \cite{DIMN} proposed recently is trained on 5 large datasets (CUHK02 \cite{CUHK02}, CUHK03 \cite{CUHK03}, Market-1501 \cite{Market-1501}, DukeMTMC-reID \cite{DukeMTMC-reID}, and CUHK-SYSU \cite{CUHK-SYSU}) and tested on 4 small benchmarks (VIPeR \cite{VIPeR}, PRID \cite{PRID}, GRID \cite{GRID},and i-LIDS \cite{i-LIDS}). The DIMN method \cite{DIMN} follows the meta-learning approach \cite{PPA}. Different from the common way of using feature distances for matching scores, DIMN generates classifier weights from gallery images and then takes the inner product of the classifier weights and probe image features to calculate matching scores. This meta-learning pipeline makes the model domain-invariant, but the complicated meta-learning procedure makes optimization difficult. In addition, classifier weight generation during testing slows down the speed of model inference. Considering these drawbacks, a simpler approach, DualNorm  \cite{DualNorm}, that utilizes normalization was proposed. Unlike DIMN, the DualNorm method focuses on learning domain-invariant features. It regards the style and content variations as the cause of domain bias and suppresses them by inserting instance normalization (IN) \cite{IN} in the early layers and a batch normalization (BN) \cite{BN} to a feature extraction layer. Both DIMN and DualNorm only use the person identity labels during the model training. They do not fully utilize domain labels (dataset labels). In our proposed MMFA-AAE method, we use the MMD-based \cite{MMD} adversarial domain learning to suppress the domain-specific information.

\section{The Proposed Methodology}
\noindent\textbf{Domain Aggregation Baseline}\\
To evaluate the effectiveness of the proposed MMFA-AAE model, we first build a Person Re-ID model to serve as the baseline reference. This baseline model uses MobileNetV2 \cite{MobilenetV2} and ResNet50 \cite{ResNet} as the backbone. We keep the default structure of the backbone and only change the dimension of the last classification layer (fully connected layer) to the total number of identities. Similar to \cite{DIMN,DualNorm}, the baseline model is trained on labeled images aggregated from multiple source domains. Let $\mathbf{X} = \left [ \mathbf{x}_{1},...,\mathbf{x}_{n} \right ]$ be the extracted feature vectors (feature embeddings) from the backbone network with batch size $n$ and $\mathbf{Y} = \left [ \mathbf{y}_{1},...,\mathbf{y}_{n} \right ]$ the corresponding person identity label set of $\mathbf{X}$. The mini-batch contains samples randomly selected from all source domains. The baseline model is pre-trained on ImageNet \cite{ImageNet} and jointly optimized with the cross-entropy loss $\mathcal{L}_{id}$ for identity classification and the triplet loss $\mathcal{L}_{tri}$ for people verification. $p_{id}(\mathbf{x}_{i},\mathbf{y}_{i})$ denotes the predicted probability that feature vector $\mathbf{x}_{i}$ belongs to person identity $\mathbf{y}_{i}$. The identity loss $\mathcal{L}_{id}$ can be expressed as
\begin{align}
\mathcal{L}_{id} = \frac{1}{n}\sum^{n}_{i=1}log(p_{id}(\mathbf{x}_{i},\mathbf{y}_{i})).
\end{align}

In every mini-batch, the images can be divided into three groups, anchor images, positive pairs to the anchor and negative pairs to the anchor. The feature embeddings $\mathbf{X}_{a}$ of the images of a person are used as the \textit{anchor} of the triplet. $\mathbf{X}_{p}$ denotes the different feature embeddings of the same person of the anchor image (positive pairs to the anchor image). $\mathbf{X}_n$ denotes the feature embeddings of different people (negative pairs to the anchor image). The training process encourages the model to make the $l_2$ distance between the positive pair $d_{ap}=d(\mathbf{X}_a,\mathbf{X}_p)$ smaller than the negative pair $d_{an}=d(\mathbf{X}_a,\mathbf{X}_n)$ by a margin $\alpha_1$. The triplet loss function $\mathcal{L}_{tri}$ of one triplet can be defined as
\begin{equation}
\begin{aligned}
\mathcal{L}_{tri}&=\max \left\{0,d_{ap}-d_{an}+\alpha_1\right\}\\
&=\max \left\{0,d(\mathbf{X}_a,\mathbf{X}_p)-d(\mathbf{X}_a,\mathbf{X}_n)+\alpha_1\right\}.
\end{aligned}
\end{equation}

Our baseline model follows the same settings of most triplet-based models with the distance margin $\alpha$ set to 0.3. The overall loss for the baseline $\mathcal{L}_{baseline}$ will be the summation of the cross-entropy loss and the triplet loss:
\begin{align}
\mathcal{L}_{baseline} = \mathcal{L}_{id} + \mathcal{L}_{tri}
\end{align}

We use the Euclidean distance of the extracted feature vectors from the baseline network to perform person retrieval evaluation. The performance of the baseline model (ResNet50) on a single dataset setting is shown in Table \ref{tab:performance_degradation}. \\

% The cross-dataset performance on other Person Re-ID benchmarks is reported in Table \ref{tab:result}.

\noindent\textbf{MMFA-AAE Network}\\
Most domain generalization methods assume that there exists a common feature space that is able to span both seen source domains and unseen target domains. If a model can extract features from this common feature space, it is able to generalize well to other unseen domains. In order to find this feature space, we extend our previous work \textbf{M}ulti-task \textbf{M}id-level \textbf{F}eature \textbf{A}lignment network (MMFA) with an additional \textbf{A}dversarial \textbf{A}uto-\textbf{E}ncoder (AAE) \cite{AAE} to the multi-domain setting. We call it \textbf{MMFA} with \textbf{A}dversarial \textbf{A}uto-\textbf{E}ncoder (MMFA-AAE). The proposed method aims to learn a model from multiple labeled datasets and removes the domain-specific information via domain-based adversarial learning. The proposed network also minimizes the mid-level feature distribution variance based on the MMD distance \cite{MMD}. In this section, we describe how the proposed MMFA-AAE network is designed for domain generalization.

\subsection{Architecture}
\begin{figure*}[ht]
\centering
\includegraphics[width=1\textwidth]{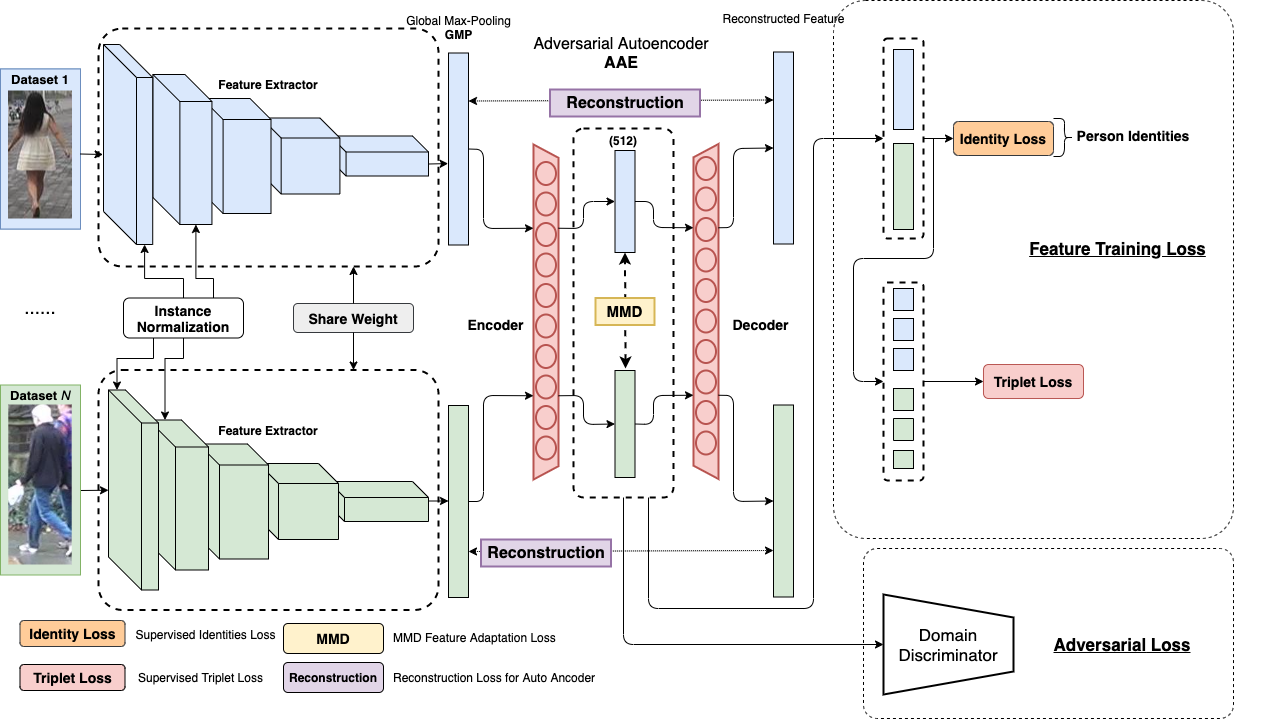}
\caption{An overview of MMFA-AAE framework for Person Re-ID multi-domain generalization. Best view in color.}
\label{fig_c6:architecture}
\end{figure*}

% The global max-pooling layer will select the maximum value from every feature maps and form a 1024 (MobileNetV2) or 2048 (ResNet50) dimension feature vectors.

The architecture of the proposed MMFA-AAE network is shown in Figure \ref{fig_c6:architecture}. In this model, we add several components to the baseline proposed at the beginning of this section. The images from multiple domains will be the inputs for the same backbone networks (MobileNetV2 \cite{MobilenetV2} or ResNet50 \cite{ResNet}) with shared weights. The feature vectors extracted from the backbone network will then be passed on to an adversarial auto-encoder \cite{AAE}. The auto-encoder \cite{AE} aims to map the feature vectors $\mathbf{X}$ from different datasets to a common latent space (hidden codes $\mathbf{H}$). The hidden codes from the auto-encoder serve as new compressed feature vectors, which are used for supervised feature training and domain discrimination. The domain discriminator determines the dataset from which the feature vector is drawn. By using the strong domain discriminator to train the feature extractor in an adversarial manner, the MMFA-AAE network aims to produce a domain-invariant latent space among multiple domains (multiple datasets). In order to further generalize the feature representation across multiple domains, we follow our previous MMFA method \cite{MMFA}, which uses Maximum Mean Discrepancy (MMD) \cite{MMD} regularization to align the distribution of the extracted deep features between different domains. In the following section, we will describe how the proposed MMFA-AAE network generalizes the feature representation from multiple domains.

\subsection{Instant Normalization}
From recent studies on the generative adversarial network (GAN), especially in the style transformation area \cite{BIN,IBN-Net}, it is observed that some image style information can be encoded in the mean and variance of the convolutional feature maps inside the network \cite{BIN}. Hence, the instance normalization (IN) \cite{IN}, which performs the normalization on a single image across all channels, can potentially eliminate the appearance divergence caused by style variation \cite{IBN-Net}. Therefore, the IBN-Net was proposed to enhance the generalization capability of the network for various computer vision tasks \cite{IBN-Net}. The DualNorm method \cite{DualNorm} applied this technique to the Person Re-ID problem and boosted the identification performance in the multi-dataset domain generalization setting. Hence, our MMFA-AAE network adopts the same setting as in \cite{DualNorm} and applies the IN in the first 6 blocks in MobileNetV2 and the first 4 blocks in ResNet50.

\subsection{MMD-Regularized Adversarial Auto-Encoder}
\subsubsection{Reconstruction Loss}
In the domain adversarial auto-encoder of our MMFA-AAE network, we use an encoder $Q(\mathbf{X})$ to map the feature embeddings $\mathbf{X}$ to the hidden codes $\mathbf{H}$ (i.e., $\mathbf{H}=Q\left(\mathbf{X}\right)$ ) and a decoder $P(\mathbf{H})$ to reconstruct the feature embeddings $\mathbf{X}$ from the hidden codes $\mathbf{H}$. The encoder-decoder pair is shared across all the domains. The reconstructed feature embedding is denoted as $P\left(Q\left(\mathbf{X}\right)\right)$ and the reconstruction loss of the auto-encoder is defined as

\begin{equation} \label{eqt:rec}
\mathcal{L}_{\mathrm{rec}}=\|\mathbf{X}-P(Q(\mathbf{X}))\|_{2}^{2}
\end{equation}

\subsubsection{Identity Loss}
In our MMFA-AAE network, $\mathbf{H} = \left [ \mathbf{h}_{1},...,\mathbf{h}_{n} \right ]$ is a set of corresponding hidden codes of $\mathbf{X}$. , serves as new compressed feature vectors for supervised feature training. Hence, the identity loss $\mathcal{L}_{id}$ for the network (cf. Eq. 1) can be expressed as: 
\begin{equation} \label{eqt:id}
\begin{aligned}
\mathcal{L}_{id} &= \frac{1}{n}\sum^{n}_{i=1}log(p_{id}(\mathbf{h}_{i},\mathbf{y}_{i}))\\
&= \frac{1}{n}\sum^{n}_{i=1}log(p_{id}(Q(\mathbf{x}_{i}),\mathbf{y}_{i})).
\end{aligned}
\end{equation}

\subsubsection{Triplet Loss}
Let $\mathbf{H}_{a}$, $\mathbf{H}_{p}$, and $\mathbf{H}_{n}$ denote the hidden codes of $\mathbf{X}_{a}$, $\mathbf{X}_{p}$, and $\mathbf{X}_{n}$, respectively. The triplet verification loss $\mathcal{L}_{tri}$ for our MMFA-AAE network (cf. Eq. 2) can be expressed as
\begin{equation}\label{eqt:tri}
\begin{aligned}
\mathcal{L}_{tri}&=\max \left\{0,d(\mathbf{H}_a,\mathbf{H}_p)-d(\mathbf{H}_a,\mathbf{H}_n)+\alpha_1\right\}\\
&=\max \left\{0,d(Q(\mathbf{X}_a),Q(\mathbf{X}_p))-d(Q(\mathbf{X}_a),Q(\mathbf{X}_n))+\alpha_1\right\}
\end{aligned}
\end{equation}

\subsubsection{Adversarial Loss}
The hidden codes can create a common latent feature space for multiple domains. Although the IN helps remove the domain style information, the extracted feature vectors may still contain other kinds of domain-specific information. Hence, there is a risk that certain hidden codes could be over-fitted to the training datasets. Therefore, we impose a domain discriminator $D$ to determine the dataset from which the feature vector is to be drawn. Suppose we have $K$ different datasets in total (i.e., $K$ domains). Let  $\mathbf{Z} = \left [ \mathbf{z}_{1},...,\mathbf{z}_{n}\right ], \mathbf{z}_{i}\in\{1,2,...,K\}$ denote the corresponding domain labels of $\mathbf{X}$. Thus, the domain discriminator $D$ can be optimized by minimizing the domain classification loss defined as

\begin{equation} \label{eqt:D_loss}
\mathcal{L}_{\mathrm{D}}(D,Q)=\sum_{i=l}^{n}log(D(Q(\mathbf{x}_i),\mathbf{z}_i))
\end{equation}

\noindent where $D(\cdot)$ denotes the predicted probability that the hidden code $Q(\mathbf{x}_i)$ belongs to  domain $\mathbf{z}_i$. After the training, the domain discriminator can capture the hidden domain-specific information, which is useful for determining the source domain of the feature vector. We can then eliminate the domain information from the network via adversarial learning using the domain discriminator. The overall adversarial learning process is a mini-max optimization problem:
\begin{equation}
arg \min _{Q} \max _{D} \mathcal{L}_{\mathrm{D}}(D,Q)
\end{equation}
$Q$ can be minimized using Eq. \ref{eqt:id} and Eq. \ref{eqt:tri}. The network can learn the feature vector by mapping it to the corresponding person identity via the identity loss $\mathcal{L}_{id}$ and the triplet verification loss $\mathcal{L}_{tri}$. $D$, on the other hand, needs to be maximized in order to suppress the domain-related information. To simplify the training process, we convert the mini-max optimization problem to a full minimization process by negating the domain classification loss $\mathcal{L}_{\mathrm{D}}$. The domain adversarial loss $\mathcal{L}_{\mathrm{adv}}$ is defined as
\begin{equation}
\mathcal{L}_{\mathrm{adv}}=-\mathcal{L}_{\mathrm{D}}(D,Q).
\end{equation}
Minimizing the domain adversarial loss $\mathcal{L}_{\mathrm{adv}}$ is equivalent to maximizing the domain classification loss $\mathcal{L}_{\mathrm{D}}$. By minimizing the domain adversarial loss $\mathcal{L}_{\mathrm{adv}}$ in the feature training process, it can guide the feature extractor to produce features that are difficult for the domain discriminator to predict the corresponding domain labels. This mechanism encourages the network to focus less on the domain-specific visual information, but more on the domain-invariant features. 

\subsubsection{MMD-based Regularization}
To further enhance the domain invariance of the hidden codes, we adopt our previous MMFA architecture and incorporate the Maximum Mean Discrepancy (MMD) \cite{MMD} regularization to align the distributions among different training datasets. Let $\mathbf{H}_l = [\mathbf{h}_{l,1},\mathbf{h}_{l,2},...,\mathbf{h}_{l_{n_l}}]$ and $\mathbf{H}_t= [\mathbf{h}_{t,1},\mathbf{h}_{t,2},...,\mathbf{h}_{t,{n_t}}]$ with batch sizes $n_l$ and $n_t$ be the hidden codes extracted from the encoder of two domains, $l$ and $t$. Also let $\phi(\cdot)$ denote a mapping operation that projects the distributions onto a reproducing kernel Hilbert space (RKHS) $\mathcal{H}$ \cite{Gretton2008AProblem}. The MMD distance between domains $l$ and $t$ can be measured according to the following equation:
\begin{equation}\label{eqt:mmd}
MMD(\mathbf{H}_l,\mathbf{H}_t)^2= \left \| \frac{1}{n_l}\sum^{n_l}_{i=1}\phi(\mathbf{h}_{l,i}) - \frac{1}{n_t}\sum^{n_t}_{j=1}\phi(\mathbf{h}_{t,j}) \right \|_{\mathcal{H}}^2
\end{equation}

\noindent The arbitrary distribution of the hidden codes of different domains can be represented by using the kernel embedding technique \cite{Smola2007ADistributions,Hu2016DeepLearning}. If the kernel $k(\cdot,\cdot)$ is characteristic, the mapping to the RKHS $\mathcal{H}$ is injective \cite{Sriperumbudur2009KernelDistributions}. The injectivity indicates that the arbitrary distribution is uniquely represented by an element in RKHS. Therefore, we have a kernel function $k(\mathbf{h}_{l,i},\mathbf{h}_{t,j})=\phi(\mathbf{h}_{l,i})\phi(\mathbf{h}_{t,j})^\intercal$ induced by $\phi(\cdot)$. The MMD distance formulated in Eq. \ref{eqt:mmd} can therefore be expressed as
\begin{equation}
\begin{aligned}
MMD(\mathbf{H}_l,\mathbf{H}_t)^2=&\frac{1}{({n_l})^2}\sum^{n_l}_{i=1}\sum^{n_l}_{i'=1}k(\mathbf{h}_{l,i},\mathbf{h}_{l,{i'}})\\
&+\frac{1}{({n_t})^2}\sum^{n_t}_{j=1}\sum^{n_t}_{j'=1}k(\mathbf{h}_{t,j},\mathbf{h}_{t,{j'}})\\
&-\frac{2}{n_l\cdot n_t}\sum^{n_l}_{i=1}\sum^{n_t}_{j=1}k(\mathbf{h}_{l,i},\mathbf{h}_{t,j})
\end{aligned}
\end{equation}

\noindent We follow the same setting as that of our previous domain adaptation MMFA model \cite{MMFA}, which uses the RBF characteristic kernel with bandwidth $\alpha_2 = 1;5;10$ to compute the MMD distance:
\begin{align}
k(\mathbf{h}_{l,i},\mathbf{h}_{t,j})=exp(-\frac{1}{2\alpha_2}\left \| \mathbf{h}_{l,i} - \mathbf{h}_{t,j} \right \|^2) 
\end{align}

\noindent Since the MMFA-AAE network focuses on the feature generalization of multiple domains ($K$ domains), the overall MMD regularization term $\mathcal{L}_{MMD}$ on the hidden codes is expressed as
\begin{equation}
\mathcal{L}_{\mathrm{MMD}}\left(\mathbf{H}_{1}, \ldots, \mathbf{H}_{K}\right)=\frac{1}{K^{2}} \sum_{1 \leq i, j \leq K} \mathrm{MMD}\left(\mathbf{H}_{i}, \mathbf{H}_{j}\right)
\end{equation}

\subsection{Training Procedure}
The learning procedure of MMFA-AAE is similar to training an AAE network \cite{AAE}. Unlike AAE, which only aims to minimize the reconstruction loss, our MMFA-AAE aims to jointly minimize the losses of identification, verification (triplet), reconstruction as well as MMD regularization on hidden codes. In our MMFA-AAE, the MMD-based adversarial auto-encoder with the early layer instance normalization enhances the feature generalization among different dataset domains. However, in order to learn a robust feature representation, the network also needs to incorporate the person identity loss and triplet loss. Our MMFA-AAE network uses the same network structure as our domain aggregation baseline proposed at the beginning of Section III. We use the same equations to compute the identity loss $\mathcal{L}_{\mathrm{id}}$ and the triplet loss $\mathcal{L}_{\mathrm{tri}}$ as formulated in Eq. \ref{eqt:id} and Eq. \ref{eqt:tri}, respectively. Unlike our baseline method, the MMFA-AAE model makes use of three additional loss functions. The reconstruction loss $\mathcal{L}_{\mathrm{rec}}$ is used to preserve the content information of the feature vectors while performing latent space projection during the dimension reduction. The MMD regularization loss $\mathcal{L}_{\mathrm{MMD}}$ helps align the distribution between different domains. The adversarial loss $\mathcal{L}_{\mathrm{adv}}$ is computed according to Eq. 9. By maximizing the domain classification loss $\mathcal{L}_{\mathrm{D}}$ as defined in Eq. 7 (i.e., minimizing $\mathcal{L}_{\mathrm{adv}}$ as defined in Eq. 9), the network is guided to suppress the domain-specific information encoded in the extracted feature vectors. Similar to training other adversarial learning models, the training procedures for the MMFA-AAE model can be divided into two phases:

\begin{enumerate}
\item Freezing the feature extractor while using the feature vectors extracted from the network to train and update the domain discriminator $D$ by minimizing $\mathcal{L}_{\mathrm{D}}$. The domain discriminator $D$ aims to predict the dataset from which a feature map is extracted. The domain classification loss can be computed with Eq. \ref{eqt:D_loss}. We repeat the same process five times in a single iteration to minimize the domain classification loss for a relatively accurate domain prediction.

\item Freezing the domain discriminator $D$ while training the feature extractor using the identity loss $\mathcal{L}_{\mathrm{id}}$ and triplet loss $\mathcal{L}_{\mathrm{tri}}$ to predict the identity labels and minimize the triplet distance, respectively. Meanwhile, the reconstruction loss $\mathcal{L}_{\mathrm{rec}}$, the MMD domain distance loss $\mathcal{L}_{\mathrm{MMD}}$ and adversarial loss $\mathcal{L}_{\mathrm{adv}}$ help to remove the domain-specific information. The feature extractor training loss $\mathcal{L}$ can thus be formulated as a weighted sum of all these losses:
\begin{equation} \label{eqt:final}
\mathcal{L} = \mathcal{L}_{\mathrm{id}}+\lambda_{1}\mathcal{L}_{\mathrm{tri}}+\lambda_{2} \mathcal{L}_{\mathrm{rec}}+\lambda_{3} \mathcal{L}_{\mathrm{MMD}}+\lambda_{4} \mathcal{L}_{\mathrm{adv}}
\end{equation}
\end{enumerate}

\noindent Let $E^{*}$, $Q^{*}$, $P^{*}$ and $D^{*}$ denotes the parameters for feature extractor, encoder, decoder and domain discriminator, respectively. The overall algorithm of MMFA-AAE is illustrated in Algorithm \ref{alg_c6:MMFA-AAE}.
\begin{algorithm}[h]
\caption{MMFA-AAE Network Training}
\label{alg_c6:MMFA-AAE}
\renewcommand{\algorithmicrequire}{\textbf{Input:}}
\renewcommand{\algorithmicensure}{\textbf{Output:}}
\begin{algorithmic}[1]
\REQUIRE Multiple Dataset Domains $\mathbf{z}_1,\mathbf{z}_2,\dots,\mathbf{z}_K$
\ENSURE Learned parameters $E^{*}$, $Q^{*}$, $P^{*}$ and $D^{*}$.
\FOR{$t=1$ to max iteration}
\STATE Sample a mini-batch of images with the corresponding person identity labels $\mathbf{Y}$ and dataset labels (domain labels) $\mathbf{Z}$
\FOR{$i=1$ to 5}
\STATE Freezing $E^{*}$, $Q^{*}$, $P^{*}$ parameters and sample hidden codes $\mathbf{H}$ from the feature extractor and the encoder.
\STATE Compute the gradient of Eq. \ref{eqt:D_loss} with respect to $D$ on hidden codes $\mathbf{H}$ and the corresponding domain labels $\mathbf{Z}$.
\STATE Use the gradient to update $D^{*}$ by minimizing the objective of Eq. \ref{eqt:D_loss}.
\ENDFOR
\STATE Freezing $D^{*}$ parameters and sample hidden codes $\mathbf{H}$ from the feature extractor and the encoder.
\STATE Compute the gradient of Eq. \ref{eqt:final} with respect to $E^{*}$, $Q^{*}$, $P^{*}$ on $\mathbf{H}$ and the corresponding person identity labels $\mathbf{Y}$.
\STATE Use the gradient to update $E^{*}$, $Q^{*}$, $P^{*}$ by minimizing the objective of Eq. \ref{eqt:final}.
\ENDFOR
\end{algorithmic}
\end{algorithm}

\section{Experiments}
\subsection{Datasets and Settings}
To evaluate our method, we follow the experiment settings in the DIMN method \cite{DIMN}, which were also adopted by DualNorm \cite{DualNorm}. In these settings, multiple large-scale benchmark datasets are combined to train a model. Small-scale datasets are individually used to evaluate the domain generalization ability of the model. In the experiments, the CUHK02 \cite{CUHK02}, CUHK03 \cite{CUHK03}, Market-1501 \cite{Market-1501}, DukeMTMC-reID \cite{DukeMTMC-reID} and CUHK-SYSU \cite{CUHK-SYSU} datasets are selected for training. All these datasets have more than one thousand identities and thousands of images. We use all the images in this combined dataset to train our model, regardless of their original training/testing splits. All Person Re-ID models involved in the comparisons are trained with $121,765$ images from $18,530$ identities. The statistics of the training dataset are shown in Table \ref{tab:training_dataset}. We test the models on the VIPeR \cite{VIPeR}, PRID \cite{PRID}, GRID \cite{GRID} and i-LIDS \cite{i-LIDS} datasets. However, these datasets are relatively small and have no more than one thousand identities. To evaluate the models in a more realistic manner, we also include the currently largest dataset, MSMT17 \cite{MSMT17}, in the experiments. The overall statistics of the testing datasets are shown in Table \ref{tab:testing_dataset}. The evaluation of the performance of our domain generalization method follows the same settings as in \cite{DIMN,DualNorm}.

\begin{table}[h]
\centering
\caption{The statistics of the training datasets}
\resizebox{0.48\textwidth}{!}{%
\begin{tabular}{l|c|c}
\hline
Dataset       & Total IDs  & Total Images \\ \hline
CUHK02 \cite{CUHK02}        & 1,816  & 7.264    \\
CUHK03 \cite{CUHK03}      & 1,467  & 14,097   \\
Market-1501 \cite{Market-1501}   & 1,501  & 29,419   \\
DukeMTMC-reID \cite{DukeMTMC-reID} & 1,812  & 36,411   \\
CUHK-SYSU \cite{CUHK-SYSU}     & 11,934 & 34,574   \\ \hline
Total         & 18,530 & 121,765  \\ \hline
\end{tabular}}
\label{tab:training_dataset}
\end{table}

\begin{table}[]
\centering
\caption{The statistics of testing datasets}
\resizebox{0.48\textwidth}{!}{%
\begin{tabular}{l|c|c|c|c}
\hline
\multirow{2}{*}{Dataset} & \multicolumn{2}{c|}{\#Test IDs} & \multicolumn{2}{c}{\# Test Images} \\ \cline{2-5} 
                         & Probe         & Gallery         & Probe           & Gallery           \\ \hline
VIPeR \cite{VIPeR}                    & 316           & 316             & 316             & 316               \\
PRID \cite{PRID}                     & 100           & 649             & 100             & 649               \\
GRID \cite{GRID}                     & 125           & 900             & 125             & 1025              \\
i-LIDS \cite{i-LIDS}                   & 60            & 60              & 60              & 60                \\ 
MSMT17 \cite{MSMT17}                   & 3,060         & 3,060           & 9,716           & 82,161              \\ \hline
\end{tabular}}
\label{tab:testing_dataset}
\end{table}

\subsubsection{Evaluation Protocols}
We follow the evaluation protocols used in \cite{VIPeR} for VIPeR, \cite{PRID} for PRID, \cite{GRID} for GRID, and \cite{i-LIDS} for i-LIDS. Because we have to use the same testing in order to compare to other methods, we randomly select half of the VIPeR dataset for testing. For the PRID dataset, we follow the same single-shot experiments as in \cite{DNS}. Since the VIPeR and PRID datasets contain only two images per person, the mean average precision (mAP) metric cannot be used. On GRID, we follow the standard testing split recommended in \cite{GRID}. On i-LIDS, two images per identity are randomly selected as the probe image and the gallery image, respectively. For all the testing datasets mentioned above, the average results over 10 random selection of the testing sets are reported. The MSMT17 dataset has already been split into training, query, and gallery set. We follow the single-query retrieval setting for the MSMT17 dataset evaluation.

The cumulative matching characteristics (CMC) curve is used for our performance evaluation, as it is the most common metric used for evaluating Person Re-ID performance. This metric is adopted since Person Re-ID is intuitively posed as a ranking problem, where each image in the gallery is ranked based on its comparison to the probe. The probability that the correct match in the ranking equal to or less than a particular value is plotted against the size of the gallery set \cite{VIPeR}. To make the comparison concise, we simplify the CMC curve by only comparing Rank 1, Rank 5, and Rank 10 successful retrieval rates. The CMC curve evaluation is valid when only one ground truth matches each given query image. The MSMT17 dataset contains multiple ground-truth images for the same person. Therefore, we use the mean average precision (mAP) proposed in \cite{Market-1501} as an additional new evaluation metric. For each query image, the average precision (AP) is calculated as the area under its precision-recall curve. The mean value of the average precision (mAP) will reflect the overall recall of the person Re-ID algorithms.

\subsubsection{Implementation Details}
For the auto-encoder sub-network, we follow the same setting as that reported in \cite{Ghifary2015DomainAutoencoders}, which uses a single hidden layer with a size of 512 neurons. The value of the hidden layer is used as an input for both the adversarial and classification sub-networks. Both sub-networks are composed of two fully-connected (FC) layers. The size of one FC layer is set to the same size as the hidden layer and while the size of the other is made the same as the identity labels. The weights for the identity and triplet losses are made equal, \textit{i.e}, $\lambda_{1}=1$. Through various testings, it is observed that the parameters $\lambda_{2}=10,\lambda_{3}=0.2,\lambda_{4}=0.5$ yield the best performance. The Adam optimizer \cite{Kingma2015Adam:Optimization} is used for all experiments. The initial learning rate is set to 0.00035 with the warm-up training technique \cite{Goyal2017AccurateHour} and is decreased by 10\% at the 40th epoch and 70th epoch, respectively. Totally, there are 120 training epochs with a batch size of 64. We implement our model in PyTorch and train it on a single Titan X GPU. The extracted features are $l_2$ normalized before matching scores are calculated.

\begin{table*}[t]
\centering
\caption{Comparison against state-of-the-art methods. (R: Rank, S: Supervised training with a target dataset, DA: Domain Adaptation, DG: Domain Generalization, -: No report)}
\resizebox{1\textwidth}{!}{%
\begin{tabular}{c|l|ccc|ccc|ccc|ccc}
\hline
\multirow{2}{*}{Type} & \multirow{2}{*}{Method} & \multicolumn{3}{c|}{VIPeR} & \multicolumn{3}{c|}{PRID} & \multicolumn{3}{c|}{GRID} & \multicolumn{3}{c|}{i-LIDS} \\ \cline{3-14} 
 &  & R 1 & R 5 & R 10 & R 1 & R 5 & R 10 & R 1 & R 5 & R 10 & R 1 & R 5 & R 10 \\ \hline
S & Ensemble \cite{Ensemble} & 45.9 & 77.5 & 88.9 & 17.9 & 40.0 & 50.0 & - & - & - & 50.3 & 72.0 & 82.5 \\
S & DNS \cite{DNS} & 42.3 & 71.5 & 82.9 & 29.8 & 52.9 & 66.0 & - & - & - & - & - & - \\
S & ImpTrpLoss \cite{ImpTripet} & 47.8 & 74.4 & 84.8 & 22.0 & - & 47.0 & - & - & - & 60.4 & 82.7 & 90.7 \\
S & GOG \cite{GOG} & 49.7 & 79.7 & 88.7 & - & - & - & 24.7 & 47.0 & 58.4 & - & - & - \\
S & MTDnet \cite{MTDnet} & 47.5 & 73.1 & 82.6 & 32.0 & 51.0 & 62.0 & - & - & - & 58.4 & 80.4 & 87.3 \\
S & OneShot \cite{OneShot} & 34.3 & - & - & 41.4 & - & - & - & - & - & 51.2 & - & - \\
S & SpindleNet \cite{SpindleNet} & 53.8 & 74.1 & 83.2 & 67.0 & 89.0 & 89.0 & - & - & - & 66.3 & 86.6 & 91.8 \\
S & SSM \cite{SSM} & 53.7 & - & 91.5 & - & - & - & 27.2 & - & 61.2 & - & - & - \\
S & JLML \cite{JLML} & 50.2 & 74.2 & 84.3 & - & - & - & 37.5 & 61.4 & 69.4 & - & - & - \\ \hline
DA & MMFA(Market-1501) \cite{MMFA} & 39.1 & - & - & 35.1 & - & - & - & - & - & - & - & - \\
DA & MMFA(DukeMTMC-reID) \cite{MMFA} & 36.3 & - & - & 34.5 & - & - & - & - & - & - & - & - \\
DA & TJ-AIDL(Market-1501) \cite{TJ-AIDL} & 38.5 & - & - & 26.8 & - & - & - & - & - & - & - & - \\
DA & TJ-AIDL(DukeMTMC-reID) \cite{TJ-AIDL} & 35.1 & - & - & 34.8 & - & - & - & - & - & - & - & - \\
DA & SyRI \cite{SyRI} & 43.0 & - & - & 43.0 & - & - & - & - & - & 56.5 & - & - \\ \hline
DG & AGG(DIMN) & 42.9 & 61.3 & 68.9 & 38.9 & 63.5 & 75.0 & 29.7 & 51.1 & 60.2 & 69.2 & 84.2 & 88.8 \\
DG & AGG(DualNorm) & 42.1 & - & - & 27.2 & - & - & 28.6 & - & - & 66.3 & - & - \\
DG & AGG(MMFA-AAE) & 48.1 & - & - & 27.7 & - & - & 32.6 & - & - & 67.3 & - & - \\
DG & DIMN \cite{DIMN} & 51.2 & 70.2 & 76.0 & 39.2 & 67.0 & 76.7 & 29.3 & 53.3 & 65.8 & 70.2 & 89.7 & 94.5 \\
DG & DualNorm \cite{DualNorm} & \underline{53.9} & - & - & \textbf{60.4} & - & - & \underline{41.4} & - & - & \underline{74.8} & - & - \\
DG & \textbf{MMFA-AAE} & \textbf{58.4} & - & - & \underline{57.2} & - & - & \textbf{47.4} & - & - & \textbf{84.8} & - & - \\ \hline
\end{tabular}}
\label{tab:result}
\end{table*}

\subsection{Comparison against state-of-the-art methods}
To demonstrate the superiority of our method, we compare it with various state-of-the-art methods under three different experimental conditions: fully supervised, unsupervised domain adaptation, and domain generalization. In Table \ref{tab:result}, the \textit{DG} methods are the multi-dataset domain generalization approaches. The AGG methods in the \textit{DG} category are the domain aggregation baselines trained without any domain generalization layer or sub-network. \textit{S} denotes a fully supervised method trained using images and labels from the corresponding target dataset. The \textit{DA} methods utilize unsupervised domain adaptation techniques. It is important to note that the \textit{DA} and \textit{S} methods are advantaged in the comparison in the sense that they have information about the target domain while our MMFA-AAE does not. We include them not as direct competitors, but to contextualize our results.

\subsubsection{Comparison with Domain Generalization Methods} 
As discussed earlier, domain generalization (DG) is the most practical approach to the Person Re-ID problem. It assumes that a target dataset cannot be seen during training. Because of this challenge, domain generalization methods have to learn a domain-invariant feature representation from other datasets. However, there are only few prior studies \cite{DIMN,DualNorm} on domain generalization for the Person Re-ID task. To make a fair comparison with these methods, we use the same MobileNetV2 \cite{MobilenetV2} feature extractor backbone and follow the same evaluation protocol and experiment settings as those adopted in \cite{DIMN} and \cite{DualNorm}. The lower part of Table \ref{tab:result} shows the benchmark results of the methods. Our AGG baseline is slightly higher because of the additional triplet loss used during the supervised training. The MMFA-AAE network attains a 10\% to 30\% increase in terms of Rank 1 retrieval accuracy for all four datasets. Our MMFA-AAE method outperforms the DIMN and DualNorm on VIPeR, GRID and i-LIDS by a large margin. MMFA-AAE only falls behind DualNorm by 3\% in Rank 1 accuracy when tested on the PRID dataset but still performs nearly 20\% higher than the DIMN method.

\begin{table}[h]
\centering
\caption{Comparison between DualNorm and MMFA-AAE with ResNet50 backbone on the MSMT17 dataset}
\resizebox{0.48\textwidth}{!}{%
\begin{tabular}{l|c|c|c|c}
\hline
\multicolumn{1}{c|}{\multirow{2}{*}{Model}} & \multicolumn{4}{c}{MSMT17} \\ \cline{2-5} 
\multicolumn{1}{c|}{} & Rank 1 & Rank 5 & Rank 10 & mAP \\ \hline
AGG(MMFA-AAE) & 14.8 & 27.8 & 37.6 & 5.9 \\
DualNorm & 42.6 & 55.9 & 61.8 & 19.6 \\
\textbf{MMFA-AAE} & \textbf{46.0} & \textbf{59.5} & \textbf{64.2} & \textbf{20.7} \\ \hline
\end{tabular}}
\label{tab:result_MSMT17}
\end{table}

To further demonstrate the proposed MMFA-AAE's superiority to other methods, we also conduct the experiments on the largest Person Re-ID benchmark: MSMT17. Table \ref{tab:result_MSMT17} provides a performance comparison of our domain aggregation baseline, the DualNorm method and our MMFA-AAE network. All three methods use the same ResNet50 backbone to allow a fair comparison. The domain aggregation baseline without any domain generalization capability can only achieve 14.8\% Rank 1 accuracy and 5.9\% mAP score. Both DualNorm and our MMFA-AAE outperform the baseline method by a large margin in both Rank 1 and mAP scores. Our MMFA-AAE consistently surpasses the DualNorm by 3 to 4\% in terms of Rank 1, Rank 5, and Rank 10 accuracy. Overall, our MMFA-AAE yields a much better performance most of the time without any additional data collection and domain adaptation process.

\subsubsection{Comparison with Domain Adaptation Methods}
We also compare our MMFA-AAE with other unsupervised domain adaptation methods. Multi-dataset domain generalization approaches focus on learning the universal feature representation from multiple datasets and assume the model can learn well-generalized features for any unseen camera network. Domain adaptation (DA) approaches focus on analyzing the characteristics between the images from labeled datasets and unlabeled images obtained from the new camera systems. Note, as discussed earlier, the DA methods' requirement for the unlabeled images from the new camera systems makes them impractical. Although the training and experimentation setting is different for DA and DG models, our MMFA-AAE model without using any target domain image surpasses the latest unsupervised domain adaptation approaches such as TJ-AIDL \cite{TJ-AIDL}, MMFA \cite{MMFA}, and SyRI \cite{SyRI}. The performance of the DA methods is shown in the middle section of Table \ref{tab:result}. MMFA-AAE outperforms all of these DA methods on all the benchmark datasets without using any image from the target dataset and does not use additional adaptation. This means that our method can effectively use the domain-invariant feature learned from multiple large-scale datasets.

\subsubsection{Comparison with Supervised Methods}
Although many fully supervised methods are reported to have high performance on large-scale datasets such as Market-1501 and DukeMTMC-reID, their performance is still low when trained on small-scale datasets. Many methods have been proposed to address this issue \cite{Ensemble,DNS,ImpTripet,GOG,MTDnet,OneShot,SpindleNet,SSM,JLML}. We have selected several supervised methods (labeled as \textit{S} in Table \ref{tab:result}) with reports on at least one of the four benchmark datasets. These methods are Ensemble \cite{Ensemble}, DNS \cite{DNS}, ImpTriplet\cite{ImpTripet}, GOG \cite{GOG}, MTDnet \cite{MTDnet}, OneShot \cite{OneShot}, SpindleNet \cite{SpindleNet}, SSM \cite{SSM}, and JLML \cite{JLML}. They follow conventional single-dataset training and testing procedures. It is not a fair comparison for MMFA-AAE method, which operates under the more challenging cross-dataset generalization setting. However, we use their results as references to illustrate the generalization capability of our MMFA-AAE model. Our MMFA-AAE method shows competitive or even better results on all four benchmarks. 

Overall, our proposed MMFA-AAE network demonstrates state-of-the-art performance. It can effectively reduce the influence of domain-specific features by using the adversarial training method and learn a more general feature representation. 
\subsection{Ablation study}
\begin{table*}[h]
\centering
\caption{Ablation study on the impact of different components for MMFA-AAE networks}
\resizebox{1\textwidth}{!}{%
\begin{tabular}{l|l|l|l|l}
\hline
\multirow{2}{*}{Method} & \multicolumn{1}{c|}{VIPeR} & \multicolumn{1}{c|}{PRID} & GRID & i-LIDS \\ \cline{2-5} 
 & \multicolumn{1}{c|}{R-1} & \multicolumn{1}{c|}{R-1} & R-1 & R-1 \\ \hline
Baseline (ResNet50) & 42.9 & 38.9 & 29.7 & 69.2 \\
Baseline + IN (DualNorm) & 54.4 & 68.6 & 43.7 & 72.2 \\
Baseline + IN + Triplet & 55.9 & 61.6 & 43.0 & 74.8 \\
Baseline + IN + Triplet + AAE & 57 & \textbf{67.6} & 46.3 & 82.3 \\
Baseline + IN + Triplet + AAE + MMD (MMFA-AAE) & \textbf{58.4} & \underline{65.7} & \textbf{47.4} & \textbf{84.8} \\ \hline
\end{tabular}}
\label{tab:ablation}
\end{table*}

\begin{figure}[h]
\centering
\begin{minipage}[t]{0.38\textwidth}
\includegraphics[clip,width=0.23\textwidth]{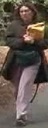}
\includegraphics[trim=6.38cm 1.3cm 6.38cm 1.3cm,clip,width=0.23\textwidth]{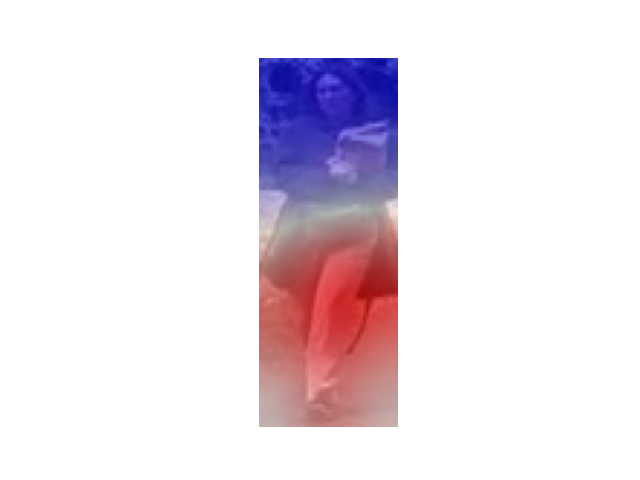}
\includegraphics[trim=6.38cm 1.3cm 6.38cm 1.3cm,clip,width=0.23\textwidth]{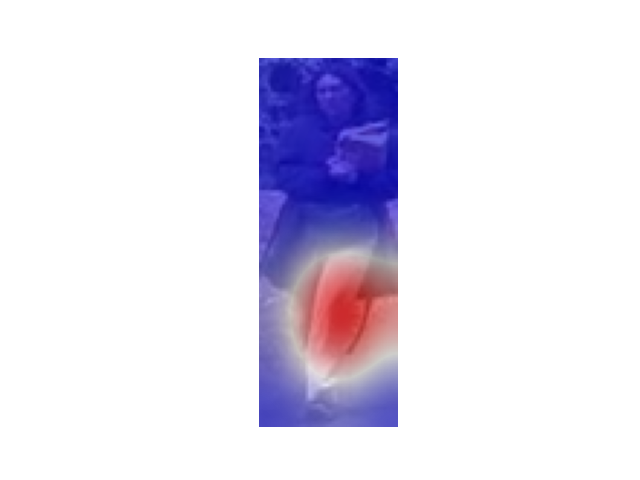}
\includegraphics[trim=6.38cm 1.3cm 6.38cm 1.3cm,clip,width=0.23\textwidth]{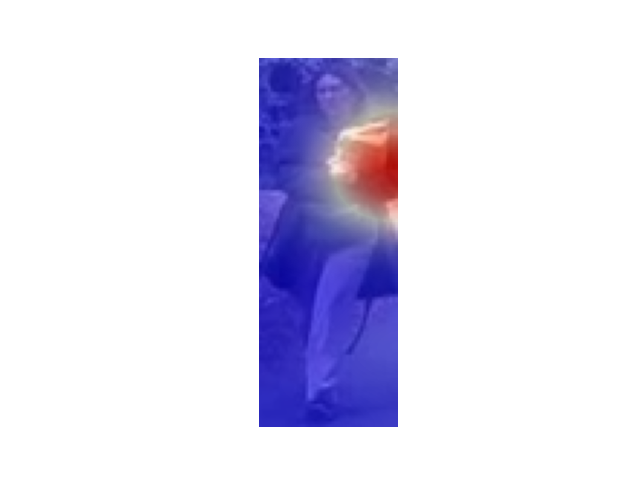}
\small \centering (a) VIPeR (Raw, Baseline, DualNorm, MMFA-AAE)
\end{minipage}
\hspace{1em}
\begin{minipage}[t]{0.38\textwidth}
\includegraphics[clip,width=0.23\textwidth]{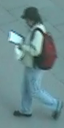}
\includegraphics[trim=5.80cm 1.3cm 5.80cm 1.3cm,clip,width=0.23\textwidth]{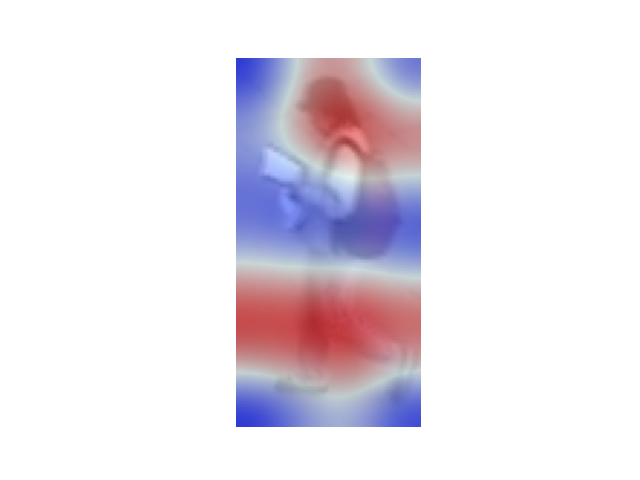}
\includegraphics[trim=5.8cm 1.3cm 5.8cm 1.3cm,clip,width=0.23\textwidth]{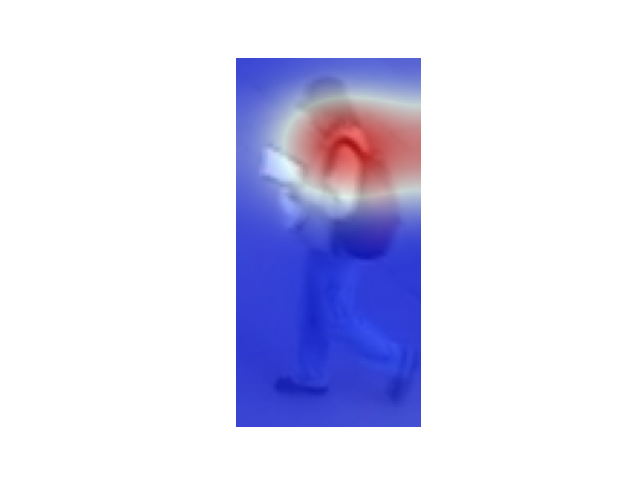}
\includegraphics[trim=5.8cm 1.3cm 5.8cm 1.3cm,clip,width=0.23\textwidth]{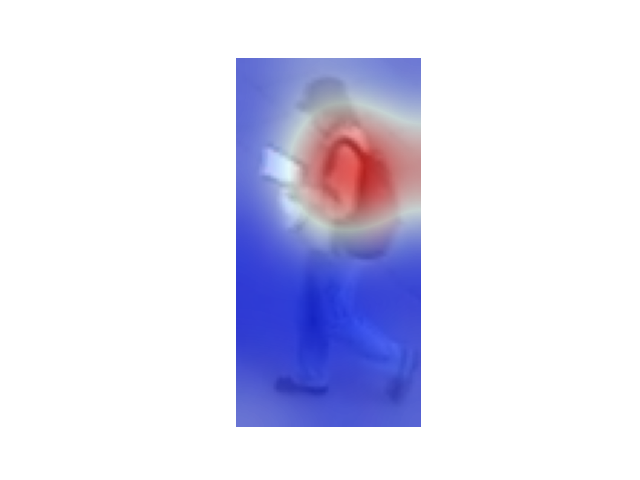}
\small \centering (b) PRID (Raw, Baseline, DualNorm, MMFA-AAE)
\end{minipage}
\hspace{1em}
\begin{minipage}[t]{0.38\textwidth}
\includegraphics[clip,width=0.23\textwidth]{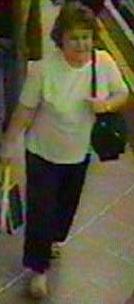}
\includegraphics[trim=6.05cm 1.3cm 6.05cm 1.3cm,clip,width=0.23\textwidth]{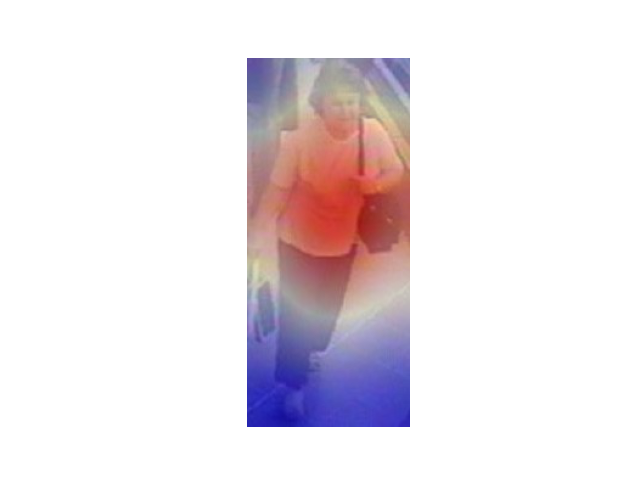}
\includegraphics[trim=6.05cm 1.3cm 6.05cm 1.3cm,clip,width=0.23\textwidth]{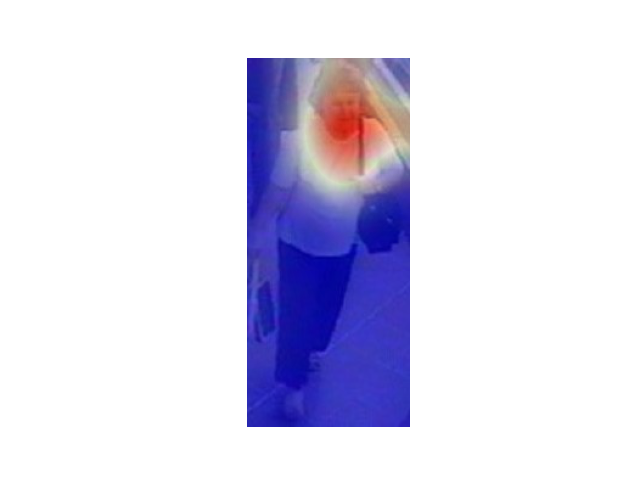}
\includegraphics[trim=6.05cm 1.3cm 6.05cm 1.3cm,clip,width=0.23\textwidth]{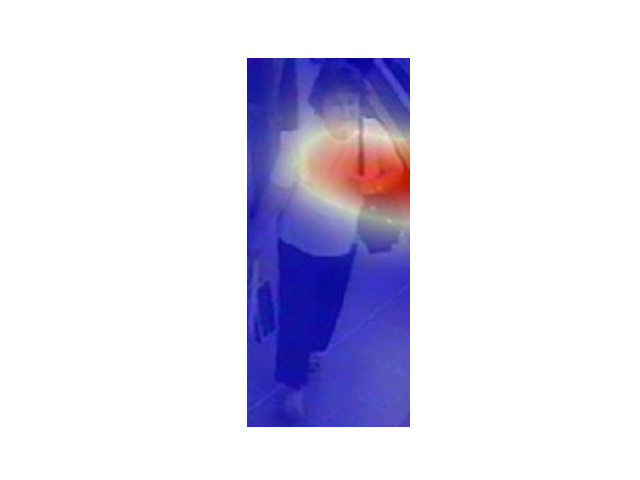}
\small \centering (c) GRID (Raw, Baseline, DualNorm, MMFA-AAE)
\end{minipage}
\hspace{1em}
\begin{minipage}[t]{0.38\textwidth}
\includegraphics[clip,width=0.23\textwidth]{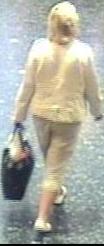}
\includegraphics[trim=6.05cm 1.3cm 6.05cm 1.3cm,clip,width=0.23\textwidth]{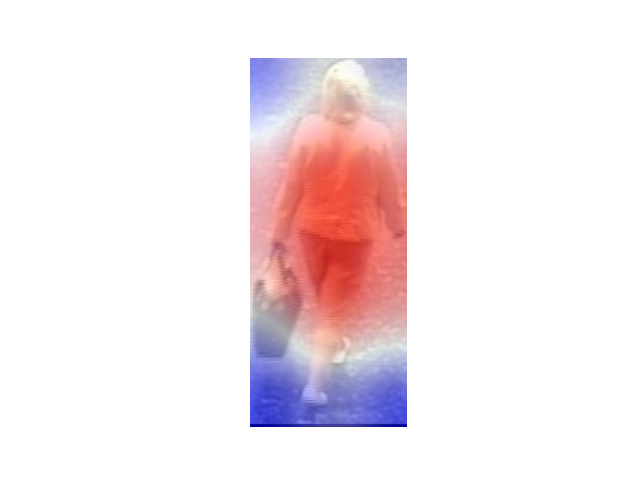}
\includegraphics[trim=6.05cm 1.3cm 6.05cm 1.3cm,clip,width=0.23\textwidth]{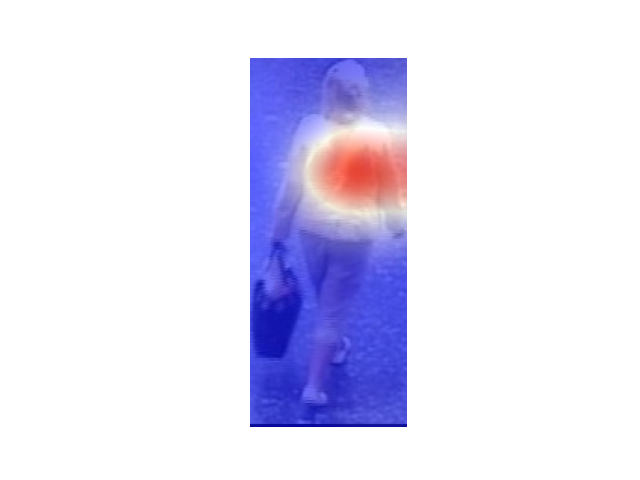}
\includegraphics[trim=6.05cm 1.3cm 6.05cm 1.3cm,clip,width=0.23\textwidth]{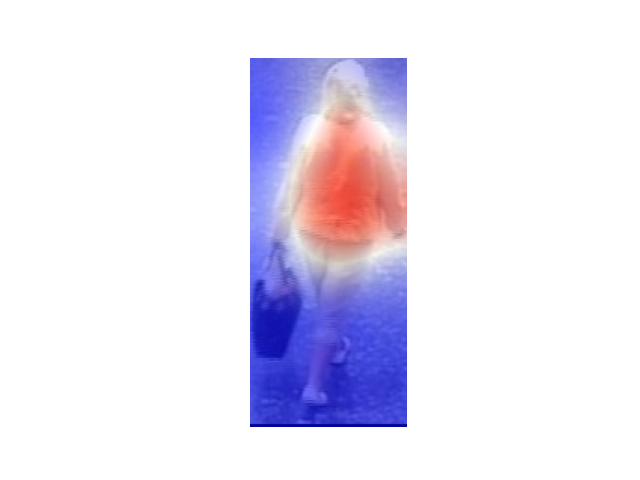}
\small \centering (c) i-LIDS (Raw, Baseline, DualNorm, MMFA-AAE)
\end{minipage}
\caption{Most activated feature maps produced by three different models on the same raw image. The images on the left-most column are the raw images while the other one shows the attention regions from the most activated feature maps of the last residual block. These feature maps highlight distinctive semantic features obtained from each model (Baseline, DualNorm and MMFA-AAE from left to right). Best view in color.}
\label{fig:feature-map}
\end{figure}

\subsubsection{Feature Heat-map Visualization}
To evaluate the effectiveness of the feature generated from our MMFA-AAE model, we randomly select images from each testing dataset and plot the most activated feature-maps obtained from the backbone network, as shown in Figure \ref{fig:feature-map}. We observed that the feature maps obtained from the domain aggregate baseline model could only focus on a vague global region. The domain generalization models such as DualNorm and MMFA-AAE can focus more on the local region with semantic meaning. In comparison with the DualNorm approach, the proposed MMFA-AAE can concentrate on the more meaningful areas like laptop or handbag, as shown in Figure \ref{fig:feature-map} (a) and (c). For images from the PRID and the i-LIDS dataset, the MMFA-AAE and DualNorm also focus on similar regions. However, the MMFA-AAE still shows superior semantic region coverage. For example, the i-LIDS image, MMFA-AAE are focusing on the entire upper torso while the DualNorm can only focus on the shoulder region.  

\subsubsection{Components Analysis}
There are four important components in the MMFA-AAE framework: Instance Normalization (IN), Triplet Loss, Adversarial Auto-Encoder (AAE), and Maximum Mean Discrepancy (MMD). To evaluate the contribution of each component, we incrementally adding one component into our baseline method and compare the performance in Table \ref{tab:ablation}. The baseline we use in the experiment uses batch normalization after global average pooling. The baseline is trained with identity loss only first. We then introduce the instance normalization into the lower convolutional layer like DualNorm. The triplet loss will further enhance the performance by 1\% to 2\% on VIPeR, GRID, and i-LIDS. The domain-based adversarial auto-encoder gives a significant 3\% to 8\% boost for all the datasets. The final MMD alignment helps further boost the overall performance by 1\% to 2\%.

\begin{figure}[h]
\centering
\begin{minipage}[t]{0.24\textwidth}
\includegraphics[clip,width=1\textwidth]{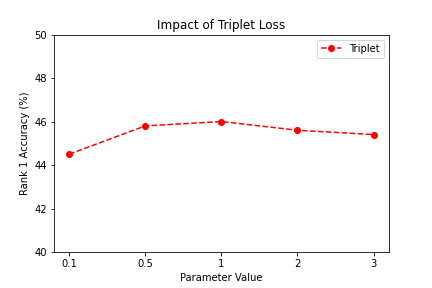}
\small \centering (a) Triplet Loss: $\lambda_{1}$
\end{minipage}
\begin{minipage}[t]{0.24\textwidth}
\includegraphics[clip,width=1\textwidth]{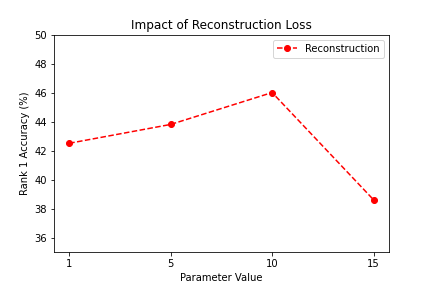}
\small \centering (b) Reconstruction Loss: $\lambda_{2}$
\end{minipage}
\hspace{1em}
\begin{minipage}[t]{0.24\textwidth}
\includegraphics[clip,width=1\textwidth]{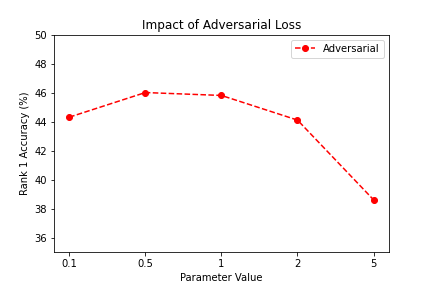}
\small \centering (c) Adversarial Loss: $\lambda_{3}$
\end{minipage}
\begin{minipage}[t]{0.24\textwidth}
\includegraphics[clip,width=1\textwidth]{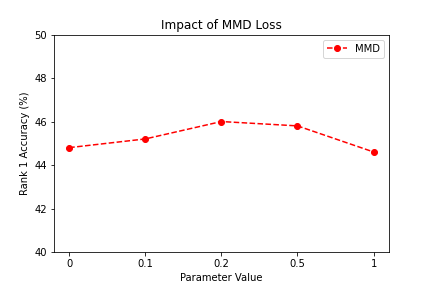}
\small \centering (d) MMD Loss: $\lambda_{4}$
\end{minipage}
\caption{The impact of the hyper-parameters of MMFA on the Re-ID Rank 1 accuracy of the MSMT17 dataset. }
\label{fig:parameters}
\end{figure}

MMFA-AAE has four hyper-parameters that affect the re-ID accuracy: $\lambda_{1}$, $\lambda_{2}$, $\lambda_{3}$ and $\lambda_{4}$. We conduct experiments to analyze the impact of these hyper-parameters. For easier comparison, we only select the largest and the most complex MSMT17 dataset for evaluation and use Rank 1 accuracy. The results are shown in Figure \ref{fig:parameters}. As shown Figure \ref{fig:parameters}, each loss function can contribute 1\% to 2\% increase to the overall performance. However, adversarial loss $\lambda_{2}$ and re-construction loss $\lambda_{3}$ need to be carefully tuned, otherwise it may even degrading the performance of the Re-ID model.

\subsubsection{t-SNE Visualization}
\begin{figure}[t] 
\centering
	\begin{subfigure}[t]{\linewidth}
		\centering
		\includegraphics[width=0.75\linewidth]{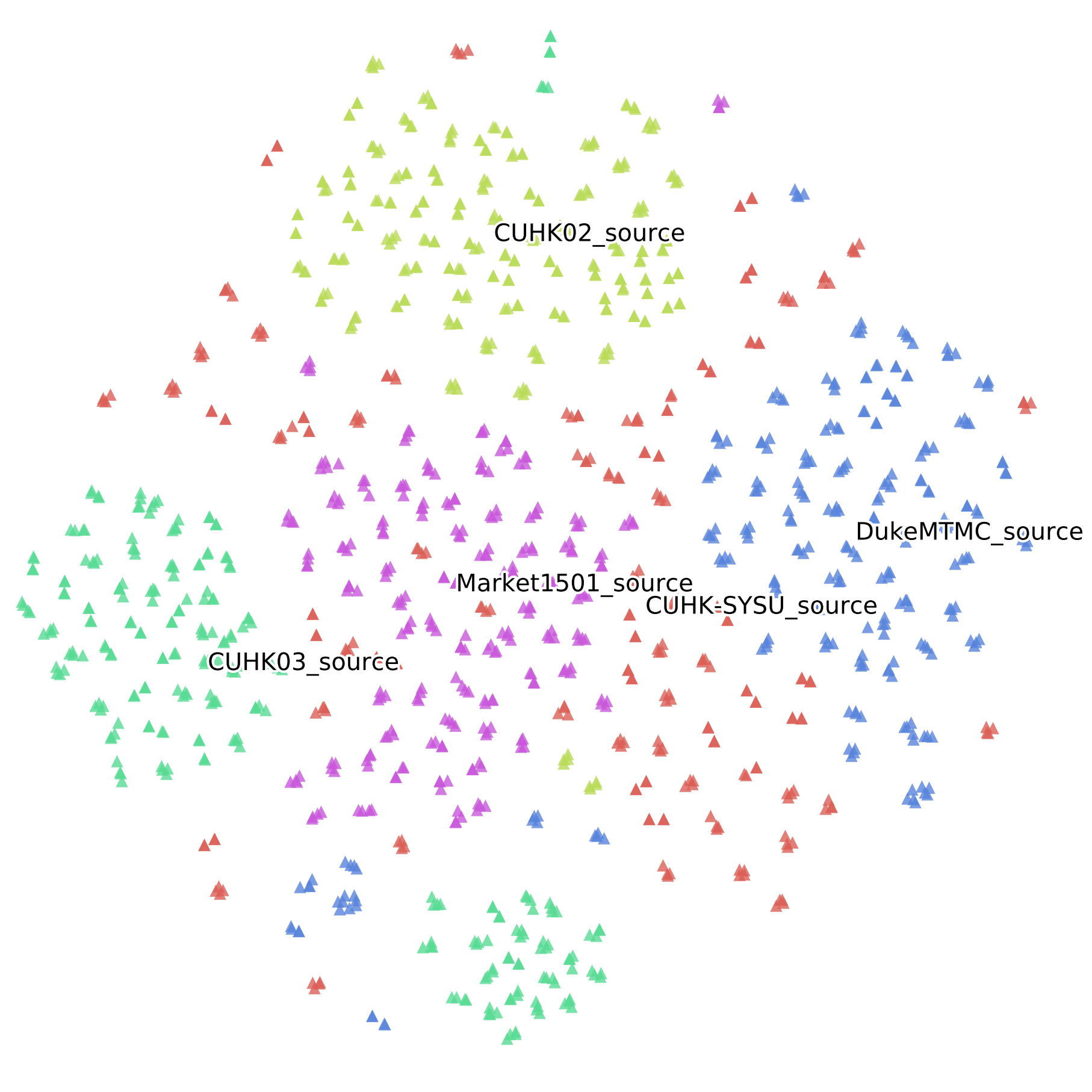}
		\caption{DualNorm}\label{fig:tsne-a}		
	\end{subfigure}
	\newline
	\begin{subfigure}[t]{\linewidth}
		\centering
	    \includegraphics[width=0.75\linewidth]{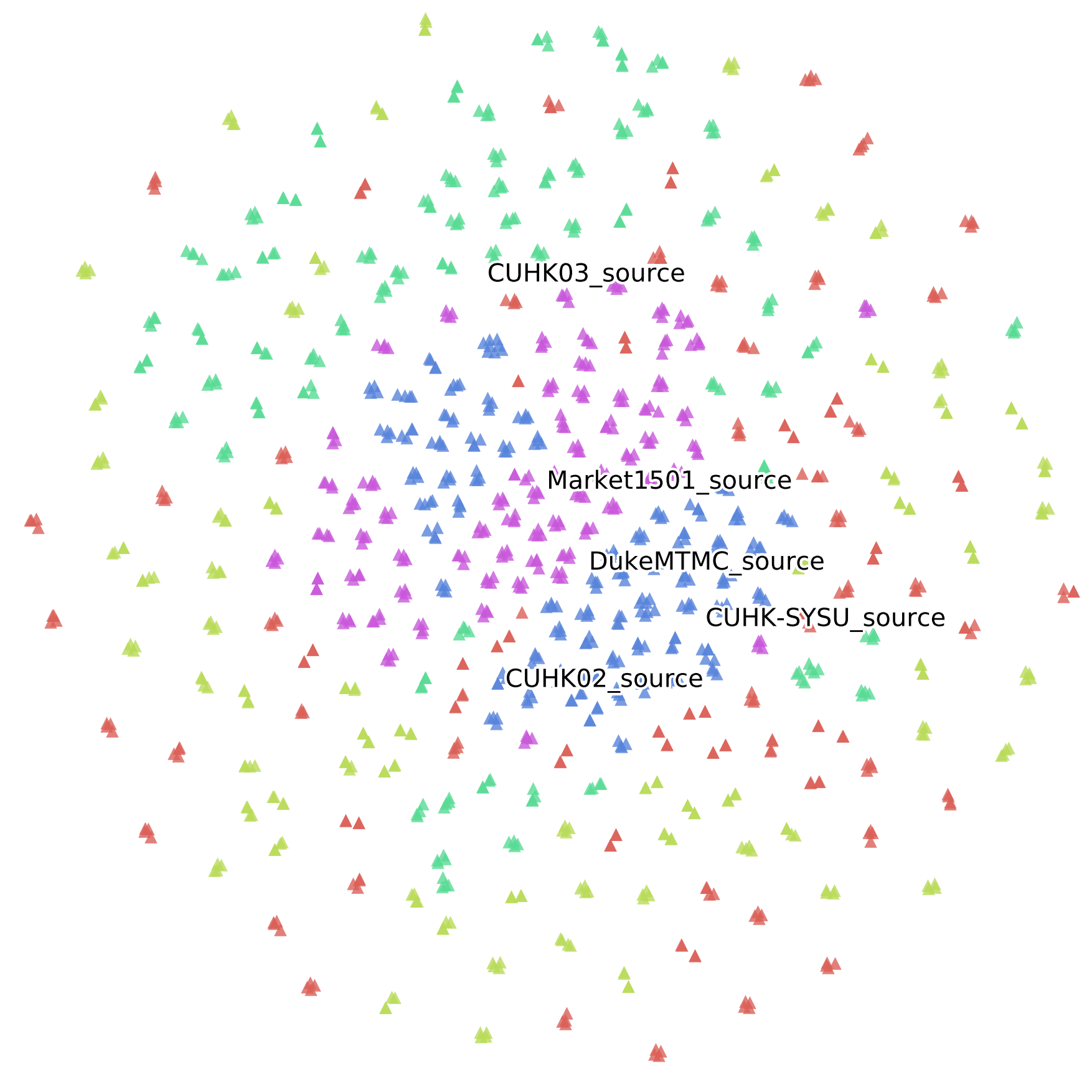}
		\caption{MMFA-AAE}\label{fig:tsne-b}
	\end{subfigure}
\caption{The t-SNE visualization of the feature vectors from the DualNorm network and our MMFA-AAE network. Different color points indicate the training dataset domains. Best view in color.}
\end{figure}

We also visualize the 2D point cloud of the feature vectors extracted from the DualNorm network and our MMFA-AAE method using t-SNE \cite{t-SNE}, as shown in Figure \ref{fig:tsne-a} and  \ref{fig:tsne-b}. We used a random sample of 6000 images from all five training datasets with a perplexity of 5000 for this visualization. As shown in Figure \ref{fig:tsne-a}, the DualNorm network can merge 5 different datasets well with low domain gaps between different datasets. However, the datasets are still clustered into several groups based on the property of the extracted feature vectors. On the other hand, our MMFA-AAE introduced the additional Adversarial-Auto-encoder (AAE) to mix up the feature vectors of different domains and alleviate the domain information. Figure \ref{fig:tsne-b} depicts our feature-point clouds extracted from the MMFA-AAE network. We can easily see that the overlap between different feature domains is more prominent in the case of the MMFA-AAE network.

\section{Conclusion}
In this paper, we propose a novel framework, MMFA-AAE, for multi-dataset feature generalization. Our MMFA-AAE network enables a Person Re-ID model to be deployed out-of-the-box for new camera networks. The main objective of our MMFA architecture is to learn a domain-invariant feature representation by jointly optimizing an adversarial auto-encoder with an MMD distance regularization. The adversarial auto-encoder is designed to learn a latent feature space among different Person Re-ID datasets via domain-based adversarial learning. The MMD-based regularization further enhances the domain-invariant features by aligning the distributions among different domains. In this way, the learned feature embedding is supposed to be universal to the seen training datasets and is expected to generalize well to unseen datasets. Extensive experiments demonstrate that our proposed MMFA-AAE is able to learn domain-invariant features, which lead to state-of-the-art performance on many datasets that it has never seen before. The proposed MMFA-AAE also out-performs most of the cross-dataset domain adaptation approaches and many fully-supervised methods. In conclusion, our MMFA-AAE approach addresses the scalability and generalization issues facing many existing Person Re-ID methods by providing a practical multi-dataset feature generalization strategy. With promising results, our MMFA-AAE approach paves the way for further research into the use of domain generalization within Person Re-ID and beyond.

\ifCLASSOPTIONcaptionsoff
  \newpage
\fi



\bibliographystyle{IEEEtran}
\bibliography{references.bib}

% \begin{IEEEbiography}[{\includegraphics[width=1in,height=1.25in,clip,keepaspectratio]{mshell}}]{Michael Shell}

\begin{IEEEbiography}[{\includegraphics[width=1in,height=1.25in,clip,keepaspectratio]{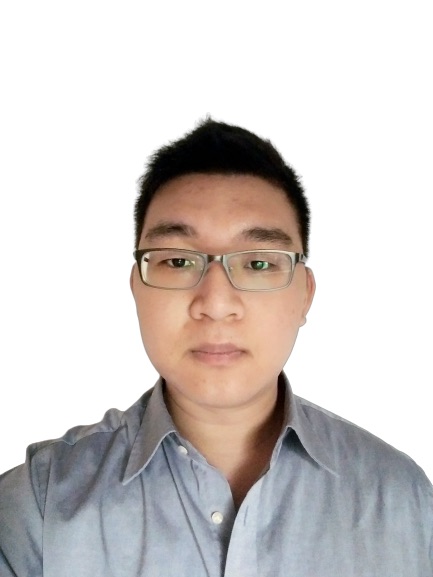}}]{Shan Lin}
received the B.Sc. degree from the University of Warwick, U.K, in 2015. He is currently pursuing the Ph.D. degree with the Department of Computer Science, University of Warwick, U.K. His current research interests are in the area of person re-identification, computer vision and deep learning. His studies are funded by the European Union EU H2020 project IDENTITY. He has published several technical papers in these areas.
\end{IEEEbiography}
\vskip -2\baselineskip plus -1fil
\begin{IEEEbiography}[{\includegraphics[width=1in,height=1.25in,clip,keepaspectratio]{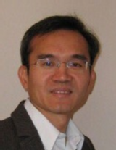}}]{Chang-Tsun Li}
received the BSc degree in electrical engineering from National Defense University, Taiwan, the MSc degree in computer science from U.S. Naval Postgraduate School, USA, and the PhD degree in computer science from the University of Warwick, UK. He is currently Professor of Cyber Security at Deakin University. He has had over 20 years research experience in multimedia forensics and security, biometrics, machine learning, data analytics, computer vision, image processing, pattern recognition, bioinformatics and content-based image retrieval. The outcomes of his research have been translated into award-winning commercial products protected by a series of international patents and have been used by a number of law enforcement agencies, national security institutions and companies around the world, including INTERPOL (Lyon, France), UK Home Office, Metropolitan Police Service (UK), Sussex Police Service (UK), Guildford Crown Court (UK), Barclays Bank PLC, US Department of Homeland Security. In addition to his active contribution to the advancement of his field of research through publication, Chang-Tsun Li is also enthusiastically serving the international cyber security community. He is currently Vice Chair of Computational Forensics Technical Committee of the International Association of Pattern Recognition (IAPR), Member of IEEE Information Forensics and Security Technical Committee, Associate Editor of IEEE Access, the EURASIP Journal of Image and Video Processing and IET Biometrics.
\end{IEEEbiography}
\vskip -2\baselineskip plus -1fil
\begin{IEEEbiography}[{\includegraphics[width=1in,height=1.25in,clip,keepaspectratio]{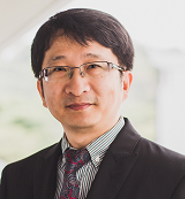}}]{Alex C. Kot}
Prof. Kot has been with the Nanyang Technological University (NTU), Singapore since 1991. He headed the Division of Information Engineering at the School of Electrical and Electronic Engineering (EEE) for eight years. He was the Vice Dean Research and Associate Chair (Research) for the School of EEE for three years, overseeing the research activities for the School with over 200 faculty members. He was the Associate Dean (Graduate Studies) for the College of Engineering (COE) for eight years. He is currently the Director of ROSE Lab [Rapid(Rich) Object SEearch Lab) and the Director of NTU-PKU Joint Research Institute . He has published extensively with over 300 technical papers in the areas of signal processing for communication, biometrics recognition, authentication, image forensics, machine learning and AI. Prof. Kot served as Associate Editor for a number of IEEE transactions, including IEEE TSP, IMM, TCSVT, TCAS-I, TCAS-II, TIP, SPM, SPL, JSTSP, JASP, TIFS, etc. He was a TC member for several IEEE Technical Committee in SPS and CASS. He has served the IEEE in various capacities such as the General Co-Chair for the 2004 IEEE International Conference on Image Processing (ICIP) and area/track chairs for several IEEE flagship conferences. He also served as the IEEE Signal Processing Society Distinguished Lecturer Program Coordinator and the Chapters Chair for IEEE Signal Processing Chapters worldwide. He received the Best Teacher of The Year Award at NTU, the Microsoft MSRA Award and as a co-author for several award papers. He was elected as the IEEE CAS Distinguished Lecturer in 2005. He was a Vice President in the Signal Processing Society and IEEE Signal Processing Society Distinguished Lecturer. He is now a Fellow of the Academy of Engineering, Singapore, a Fellow of IEEE and a Fellow of IES.
\end{IEEEbiography}
\vfill

% \begin{IEEEbiography}[{\includegraphics[width=1in,height=1.25in,clip,keepaspectratio]{photos/yingtiantang}}]{Yingtian Tang}
% is an undergraduate student in the University of Electronic Science and Technology of China. He is currently an intern at Rapid-Rich Object Search Lab, Nanyang Technological University, Singapore. His research interests include computer vision and deep learning.
% \end{IEEEbiography}


%\vfill

% \enlargethispage{-5in}

\end{document}